%% file: main.tex
\documentclass{Interspeech2024}

\interspeechcameraready
\input{00_decl}

\title{\bench: a Weakly Supervised Dataset for Hebrew Speech Processing}

\name[affiliation={1}]{Arnon}{Turetzky}
\name[affiliation={1}]{Or}{Tal}
\name[affiliation={2}]{Yael}{Segal-Feldman}
\name[affiliation={2}]{Yehoshua}{Dissen}
\name[affiliation={1}]{Ella}{Zeldes}
\name[affiliation={1}]{Amit}{Roth}
\name[affiliation={2}]{Eyal}{Cohen}
\name[affiliation={2}]{Yosi}{Shrem}
\name[affiliation={2}]{Bronya R.}{Chernyak}
\name[affiliation={2}]{Olga}{Seleznova}
\name[affiliation={2}]{Joseph}{Keshet}
\name[affiliation={1}]{Yossi}{Adi}

\address{The Hebrew University of Jerusalem}
\address{
  $^1$The Hebrew University of Jerusalem, Israel \\ 
  $^2$Technion, Israel Institute of Technology}
\email{arnon.turetzky@mail.huji.ac.il}


\begin{document}
\maketitle

\input{01_abs}
\input{02_intro}
\input{03_related}
\input{04_benchmark}

\input{05_results}
\input{06_discussion}

\newpage
\bibliographystyle{IEEEtran}
\bibliography{refs}

\end{document}

%% file: 00_decl.tex
\newcommand{\bench}{\textsc{HebDB}\xspace}
\usepackage{pifont}
\usepackage{multirow}
\usepackage{float}
\usepackage{cleveref}

\usepackage{etoolbox}
\newtoggle{release}

\togglefalse{release}
\iftoggle{release}{
\newcommand{\adios}[1]{}
\newcommand{\ort}[1]{}
\newcommand{\shua}[1]{}
\newcommand{\AT}[1]{}
}
{
\newcommand{\adios}[1]{\textcolor{red}{YA: #1}}
\newcommand{\ort}[1]{\textcolor{magenta}{OT: #1}}
\newcommand{\shua}[1]{\textcolor{green}{SD: #1}}
\newcommand{\AT}[1]{\textcolor{blue}{AT: #1}}
}

\newcommand{\newpara}[1]{\vspace{0.22cm} \noindent \textbf{#1}}

%% file: 01_abs.tex
\begin{abstract}
We present \bench, a weakly supervised dataset for spoken language processing in the Hebrew language. \bench offers roughly $2500$ hours of natural and spontaneous speech recordings in the Hebrew language, consisting of a large variety of speakers and topics. We provide raw recordings together with a pre-processed, weakly supervised, and filtered version. The goal of \bench is to further enhance research and development of spoken language processing tools for the Hebrew language. Hence, we additionally provide two baseline systems for Automatic Speech Recognition (ASR): (i) a self-supervised model; and (ii) a fully supervised model. We present the performance of these two methods optimized on \bench and compare them to current multi-lingual ASR alternatives. Results suggest the proposed method reaches better results than the evaluated baselines considering similar model sizes. Dataset, code, and models are publicly available under \url{https://pages.cs.huji.ac.il/adiyoss-lab/HebDB/}. 
\end{abstract}

%% file: 02_intro.tex
\section{Introduction}
Spoken language technologies have seen a great leap in performance following the success of deep neural networks consisting of large-scale models~\cite{wav2vec, audiopalm, wang2021unispeech} and datasets~\cite{librispeech, librilight, voxpopuli}. This includes Automatic Speech Recognition (ASR)~\cite{whisper, mms, googleusm}, Text-to-Speech (TTS)~\cite{valle, voicebox}, speech enhancement~\cite{dns, denoiser}, speaker diarizaion~\cite{ssl_shua}, to name a few. 

A fundamental requirement in the success of the aforementioned models is optimization using large-scale datasets~\cite {wav2vec}. For instance, when considering ASR, Whisper~\cite{whisper} was trained using $\sim700$k hours of speech utterances and Google USM was trained over $\sim12$M hours of speech recordings. As for TTS, both VALL-E and VoiceBox were trained over $60$k hours of speech. As a result, such big performance advancements are mainly kept for high-resource language in which large-scale datasets can be found.

One approach to mitigate the performance gaps between high-resource and low-resource languages is to train speech models considering multi-lingual setups~\cite{whisper, mms}. The authors in \cite{yadav2022survey} empirically demonstrated the benefit of training multi-lingual spoken language processing models, while the authors in \cite{mms} specifically demonstrated the benefit of low-resource languages. Although such performance improvements are interesting and important, directly training models over large-scale benchmarks still achieve superior performance~\cite{likhomanenko2020rethinking}. 

The Hebrew language is among the low-resource languages explored in prior work~\cite{whisper, mms}. The Hebrew language is being spoken by roughly 9 million people worldwide~\cite{campbell2008ethnologue}. Besides the lack of large-scale datasets in Hebrew, the language syntax and structure impose some inheriting challenges, such as: (i) using non-Latin letters, which sets it apart from many languages; (ii) Traditional Hebrew has diacritics (``Nikud'') while modern Hebrew writing rarely uses them. Such discrepancies impose a critical challenge on ASR and TTS systems which need to learn non-trivial pronunciations that directly affect word meanings. Such differences are not presented in writings but sound differently, and can be distinguished mainly based on context. For instance the word ``pita'' can have two meanings: bread and seduced based on the context; (iii) Hebrew is a morphologically rich language, with common use of prefixes and suffixes to modify words’ meanings and to add prepositions. This property makes tokenization difficult and less efficient, especially under the multi-lingual setup~\cite{petrov2024language}. 

In this study, we present \bench, a weakly supervised spontaneous speech dataset in the Hebrew language. \bench is comprised of $\sim2500$ hours of in-the-wild natural speech consisting of a numerous number of speakers and diverse topics and vocabulary. We release both the raw recordings together with a pre-processed and weakly transcribed version. We additionally provide a transcription confidence score for each of the data samples, which can be used to develop strategies for fine-tuning considering different supervision qualities. In releasing this dataset, our goal is to advance research and development of Artificial Intelligence (AI) based tools for spoken language processing directly developed for the Hebrew language. To further enhance the development of such tools, we provide two baseline systems: (i) a self-supervised model; and (ii) a fully supervised ASR model. Full dataset, code, and models are publicly available under \url{https://pages.cs.huji.ac.il/adiyoss-lab/HebDB/}.

The paper is structured as follows. We start by reviewing datasets and speech processing tools directly dedicated to the Hebrew language in \Cref{sec:related}. Next, in \Cref{sec:data} we provide a detailed description of \bench, its curation, statistics, pre-processing, and supervision quality assessment. In \Cref{sec:res}, we describe the baseline systems and compare their performance to current open-source tools. We conclude the paper in \Cref{sec:con}, where we outline future work along this research direction. 

%% file: 03_related.tex
\begin{table*}[t!]
    \centering
    \caption{Details \& Statistics of \bench's raw recordings. We report the list of sources, sampling rates, num of channels, total duration (in hours), and indication of single or multiple speakers in a given source. Both sampling rates and channels are reported percentage. \label{tab:data_stats}}
    \resizebox{1.0\textwidth}{!}{
        \begin{tabular}{lllll}
            \toprule
            \bf Source & \bf Sample rate (kHz) & \bf \# Channels & \bf Duration (h) & \bf Single / Multiple speakers \\
            \midrule
            Geekonomy               & 44.1: 60\%; 48: 40\%               & Stereo: 67\% ; Mono: 33\%   & 1146 & Multiple speakers \\
            Osim   History          & 44.1: 95\%; 48: 3\%; 22.05: 2\%    & Stereo: 93\% ; Mono: 7\%    & 270  & Mostly single speaker \\
            The Dor Kahn Experience (DKE) & 44.1: 100\%                        & Stereo: 100\%              & 218 & Multiple speakers \\            
            Yo! The   podcast       & 44.1: 25\%, 48: 75\%               & Stereo: 25\% ; Mono: 75\%  & 295 & Multiple speakers \\
            Good Question           & 16: 100\%                          & Stereo: 100\%              & 163 & Multiple speakers \\            
            Yad vashem              & 48: 100\%                          & Stereo: 100\%              & 492 & Multiple speakers\\
            \bottomrule
        \end{tabular}}
\end{table*}

\section{Related work}
\label{sec:related}
\newpara{Spoken Hebrew benchmarks.} As Hebrew is considered a low-resource language, public spoken benchmarks hardly exist. Previous efforts in constructing datasets in Hebrew were either released under a multi-lingual benchmark~\cite{black2019cmu, pratap2020mls, mms, conneau2023fleurs} or relatively small~\cite{izre2001designing, azogui2016open, marmorstein2022huji, sharoni2023saspeech}. The authors in~\cite{izre2001designing} established the Corpus of Spoken Israeli Hebrew (CoSIH) with the goal of compiling a large database of recordings of spoken Israeli Hebrew in order to facilitate and enhance research in the field. Next, the authors in~\cite{azogui2016open} released The Map Task Corpus (MaTaCOp) of Hebrew dialogues. The authors in ~\cite{marmorstein2022huji} collected naturally occurring speech and interaction in Modern Hebrew via telephone conversations during the years 2020–2021 and released the HUJI Corpus of Spoken Hebrew (HUJICorpus). More recently, the authors in ~\cite{sharoni2023saspeech} released SASPEECH, a high-quality single-speaker Hebrew dataset which is goal is to enhance Hebrew speech synthesis research. Although all of these prior work are important and valuable, the provided benchmarks are relatively small. CoSIH contains $\sim12.3$ hours of speech, the MaTaCOp corpus contains $\sim5.3$ hours, the HUJICorpus has $\sim3.8$, and SASPEECH which is the largest one contains $\sim30$ hours of speech. The most relevant concurrent work to ours is the great work done by~\cite{marmor2023ivrit}, which released a dataset denoted as \emph{ivrit.ai}. The authors released $\sim3300$ hours of speech from local podcasts and provided the first large-scale dataset in Hebrew. We would like to state that the proposed benchmark is orthogonal to the release of ivrit.ai. We believe the community should leverage as many high-quality publicly available datasets as possible to close the gap between low- to high-resource languages. Additionally, unlike ivrit.ai, we release two baseline systems (SSL and supervised one) for speech processing and ASR. 

\newpara{Hebrew ASR.} With recent advancements in multi-lingual ASR systems, we also observe improvements in Hebrew ASR. The authors in ~\cite{whisper} release the Whisper family of models that were trained on $\sim700k$ hours of labeled data from $100$ languages including Hebrew. The authors publicly released models ranging in size from $40$M to $1.55$B parameters. Later on, the authors in~\cite{mms} released the \emph{Massively Multi-lingual Speech} (MMS) project which provides speech models for $1107$ languages including Hebrew. In this work, we compare the proposed baseline systems trained on \bench to both Whisper and MMS.

%% file: 04_benchmark.tex
\section{\bench dataset}
\label{sec:data}

\bench contains natural dialogues of spontaneous speech. It is comprised of both testimonies from World War II survivors and five podcasts covering a wide range of subjects and speakers. While the testimonies provide firsthand accounts of historical events, the majority of our dataset consists of podcasts covering diverse topics such as economy, politics, sports, culture, science, history, and music, to name a few. This combination of personal narratives and informative discussions offers a rich and varied resource for analysis and interpretation.

We provide two versions of the dataset: \emph{raw} and \emph{pre-processed}. The raw version contains over 2584 hours of in-the-wild audio in varying sample rates, recorded channels, and number of speakers. \Cref{tab:data_stats} provides a detailed description of this version including statistics. We release this version to allow researchers and practitioners to explore different pre-processing alternatives and methods. 

The pre-processed version contains roughly $1690$ hours of audio, down-sampled, segmented into multiple files, and auto-transcribed. This version is better suited for training acoustic models as is. We optimize and evaluate the proposed baseline systems using the pre-processed version only. In the next sub-section, we provide a detailed description of the pre-processing pipeline used. 

Both versions of \bench corpus are released under the very permissive of CC BY $4.0$ license~\cite{creativecreative}.

\subsection{Pre-processing}
The raw recordings are constructed from full podcast episodes and testimonies and, hence, contain long audio sources and plenty of non-speech segments, e.g. music, environmental sounds, silence, etc. Such in-the-wild conditions make model optimization challenging and require a pre-processing step. 

To handle that, we apply the following pre-processing pipeline to the raw version of \bench. We first resample all the audio recordings to 16kHz, mono recordings, using julius~\footnote{\url{https://github.com/adefossez/julius}} python package. 
Next, we apply a Voice Activity Detection (VAD) model to partition the waveform to sentences and discard empty and noisy parts. Lastly, we automatically transcribe the segmented speech utterances using a pre-trained ASR model. 

\newpara{Voice activity detection and speech segmentation.} We use the \texttt{silero-vad}~\cite{silero_vad} to perform voice activity detection over the $16$KHz audio files. Unlike traditional VAD models~\cite{sohn1999statistical} that are based on Digital Signal Processing heuristics, the `silero-vad' is a learning model based on convolutional and LSTM layers~\footnote{\url{https://github.com/snakers4/silero-vad}}. We specifically chose the \texttt{silero-vad} as it provides superior performance to other publicly available VAD models and was found beneficial in prior work~\cite{marmor2023ivrit, sharoni2023saspeech}.

In general, VAD relies on frame-wise activity classification. Following that, to properly segment the audio into sentences, we need to calibrate a classification threshold over the model's frame-wise confidence scores and define a minimal duration of silence between activated segments. We follow this process as we wish to have a minimal number of words in each segment while keeping its length to fit in a processing unit memory.

Specifically, we use a confidence threshold of $0.5$ to filter out activated segments with a minimum duration of $1$ seconds, separating audio segments by a minimal silence duration of $100$ms and padding both sides of the segmented audio with $30$ms of silence.

\newpara{Transcriptions.} We provide weak supervision in the form of transcriptions. We leverage the pre-trained Whisper large-v2 (1.55B) version to transcribe all the segmented data. Although Whisper supports transcription in specific languages it might output non-Hebrew characters not limited to Latin. Analyzing the frequencies of character across our train set, most of the non-Hebrew chars were found in ~$1$\%, hence we removed those samples from our data. Additionally, Whisper might output an $<$RTL$>$ token, as this token is not relevant to the acoustics of the speech utterance we simply remove it from the text. For better alignment between acoustics and written text, we converted numbers and dates to words using the num$2$words package~\footnote{\url{https://pypi.org/project/num2words/}}. Lastly, Hebrew has $5$ special letters with final form, we experimented with normalizing the final instances to regular ones and found it to be beneficial. Hence, we adjust the $5$-gram LM provided by the MMS to normalize accordingly.

\begin{figure}[t!]
    \centering
    \includegraphics[width=\columnwidth]{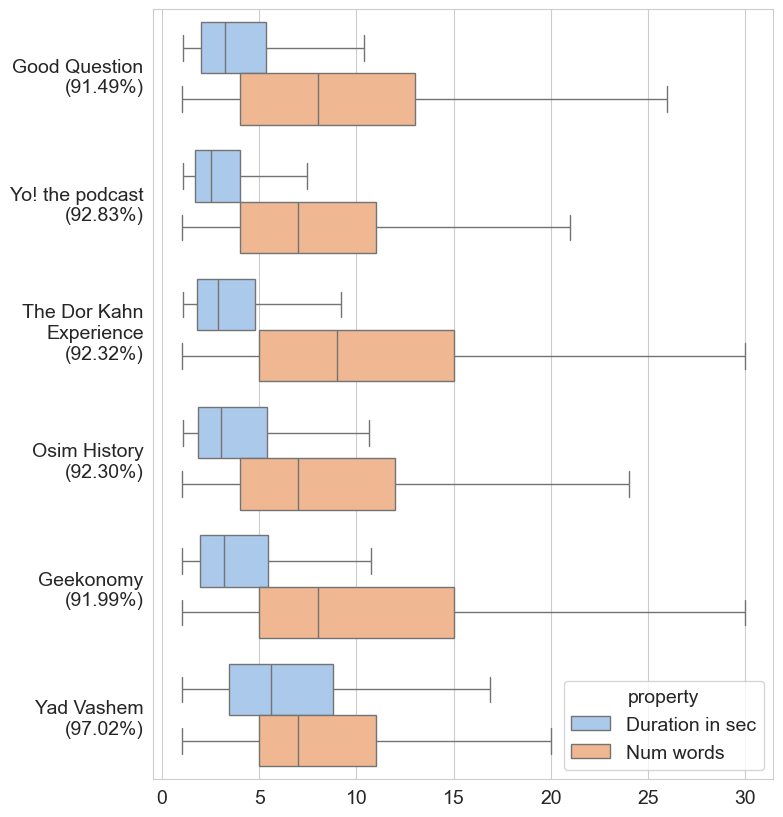}
    \caption{A boxplot of the processed data, percentages denote the corresponding portion of the unfiltered non-outlier instances with respect to each source. We provide statistics for both audio recordings (in hours) and transcriptions (in words count).}
    \label{fig:preprocessed_box}
\end{figure}

\newpara{Statistics.}
After the prepossessing step, we are left with $\sim1690$ hours of speech partitioned into varied length segments with the vast majority of the segmented files having less than $10$ seconds. \Cref{tab:processed data stat} shows a further subdivision of processed audio with respect to each source separately. ~\Cref{fig:preprocessed_box} depicts a box plot for processed instances quartile distributions over audio duration in seconds and the number of transcribed words with respect to each source, discarding outliers. 

Notice, that the pre-processing step did not affect all sources equally. For instance, the Yad vashem source was reduced from $492$ hours to $67.4$, this is due to bad recording conditions and long silences at the beginning or end of the files.

\begin{table}[t!]
    \centering
    \caption{Details of \bench's pre-processed version. We report the total duration (in hours) for each source together with statistics of the processed utterances (in seconds).}
    \label{tab:processed data stat}
    \resizebox{\columnwidth}{!}{
        \begin{tabular}{llc}
        \toprule
        \bf Source          & \bf Duration (h)  & \bf Mean / Min / Max / Med (s) \\
        \midrule
        Geekonomy       & 998.5     & 4.7 / 1.0 / 282.1 / 3.2 \\
        Osim History    & 223.9     & 4.4 / 1.1 / 152.2 / 3.1 \\
        DKE             & 181.4     & 3.8 / 1.1 / 107.6 / 2.9  \\
        Yo! the podcast & 142.2     & 3.3 / 1.1 / 115.7 / 2.5 \\
        Good Question   & 75.0      & 5.0 / 1.1 / 175.5 / 3.2 \\
        Yad Vashem      & 67.4      & 5.3 / 1.0 / 361.1 / 4.4 \\
        \bottomrule
        \end{tabular}
    }
\end{table}

\begin{figure}[t!]
    \centering
    \includegraphics[width=\columnwidth]{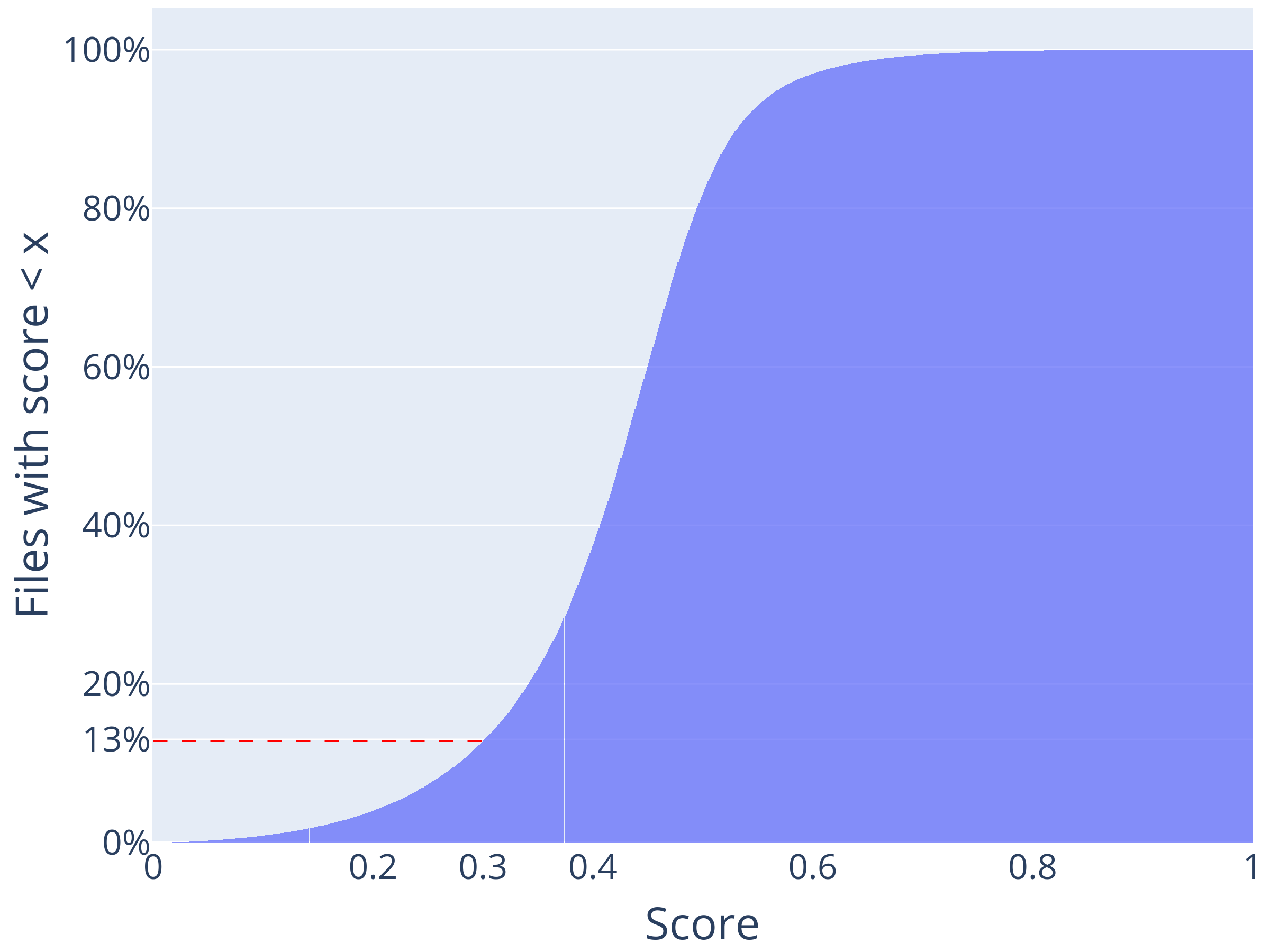}
    \caption{Score level histogram of the data filtering process. We use a threshold of 0.3 which filters roughly 13\% of the data.}
    \label{fig:df_hist}
\end{figure}

\subsection{Data filtering}
\label{sec:df}
To further enhance the reliability of our transcripts, we employ a forced aligner using an alternative model, specifically the MMS model \cite{pratap2023scaling}. This model requires the input text to be in transliterated Latin script for accurate alignment. We achieved this using the Uroman package \footnote{\url{https://github.com/isi-nlp/uroman}}, which romanizes text from most languages into the Latin alphabet. However, Uroman's performance dips with non-diacritized text, prompting us to first diacritize all transcriptions using the UNIKUD package~\footnote{\url{https://pypi.org/project/unikud/}}, a tool specifically designed to add necessary diacritical marks to Hebrew text, thus ensuring higher transliteration accuracy.
 
We utilized the forced aligner to generate a confidence score for each utterance, calculated by averaging the confidence scores of individual words. These utterance-level scores were used to filter out lower-quality data, aiming to train our models on high-quality data only. The confidence scores for Hebrew utterances were notably lower on average than for English, with a mean score of $0.417$ and a std of $0.11$. We set a threshold of $0.3$ for the confidence scores to determine the data quality cutoff. Initially, our dataset comprised $\sim1690$ hours of speech. After applying the threshold for filtering, we retained $1470$ hours of speech with a mean score of $0.447$ and a std of $0.08$, considered as reliable transcripts for training. \Cref{fig:df_hist} presents a histogram of the forced aligner scores. 

%% file: 05_results.tex
\section{Baseline system}
\label{sec:res}

\subsection{Implementation details}
We provide two baseline systems together with \bench. The first one is an SSL model, namely HuBERT~\cite{hsu2021hubert}. The second model is a fully supervised one, namely Conformer~\cite{gulati2020conformer}. Both models were optimized using \bench and evaluated on the Hebrew subset from the Fleurs benchmark~\cite{conneau2023fleurs}. Both models were evaluated with and without a language model (LM). We use $5$-gram LM provided by the MMS project~\footnote{\url{https://github.com/facebookresearch/fairseq/blob/main/examples/mms/README.md}}.

\newpara{HuBERT.} We train a HuBERT-\texttt{base} with $\sim95$M for two iterations following the standard recipe for 'pretrain' outlined in the \texttt{fairseq} framework~\cite{ott2019fairseq}\footnote{\url{https://github.com/facebookresearch/fairseq/blob/main/examples/hubert/README.md}}. In the first iteration, we utilize $100$ clusters generated from $10$\% of the data using KM-clustering on MFCC features. The model is trained on $4$ A$5000$ GPUs, using gradient accumulation to match the original recipe's specifications of $250$k training steps across $32$ GPUs. For the second iteration, we increase the number of clusters to $500$ and use representations obtained from the $6$th layer, still utilizing gradient accumulation but for $400$k training steps. After HuBERT pre-training, we employ the connectionist temporal classification (CTC) loss~\cite{graves2006connectionist} for ASR fine-tuning for $150$K steps.

\newpara{Conformer.} 
The Conformer used is based on the model introduced by Gulati et al \cite{gulati20_interspeech} trained with the CTC loss and a character tokenizer using $8$ A$40$ GPUs. The model is similar to the large model in \cite{gulati20_interspeech} with $\sim100$M parameters.
The Conformer model's hyper-parameters are as follows: convolution kernel size $31$, n-heads $8$, hidden-dim $512$, $17$ layers, and dropout $0.1$. The model is fed Mel-spectrograms of $80$ filters, with a window size of $25$ms and stride of $10$ms. 
We employ time and frequency masking as augmentation techniques during training. We utilize a Noam optimizer~\cite{vaswani2017attention} with $10,000$ warmup steps. Our batch size is measured in audio length, consisting of utterances with lengths ranging from $1$ to $30$ seconds, cumulatively not exceeding $300$ seconds per batch. Finally, we trained for a total of $160$k steps for the full training and $140$k for the filtered training.

\begin{table}[t!]
    \centering
    \caption{WER results over the Fleurs~\cite{conneau2023fleurs} benchmark. Results are reported for the provided baseline systems together with Whisper and MMS considering different setups. We provide results with and without LM and data filtering (df) whenever possible. The same LM was used in all of the reported results. \label{tab:wer}}
        \begin{tabular}{l|c|c|c}
            \toprule
            \bf Model & \bf \#param & \bf LM & \bf WER \\
            \midrule
            Whisper          & 39M    & - & 71.6 \\
            Whisper          & 74M    & - & 61.7 \\
            Whisper          & 244M   & - & 44.4 \\
            Whisper          & 769M   & - & 33.1 \\
            Whisper          & 1.55B   & - & 27.1 \\
            \midrule
            MMS (61 lang)    & 1.5B   & \ding{55}  & 66.6 \\
            MMS (61 lang)    & 1.5B   & \ding{51}  & 41.5 \\
            MMS (1,107 lang)  & 1.5B   & \ding{55}  & 67.1 \\
            MMS (1,107 lang)  & 1.5B   & \ding{51}  & 40.0 \\
            \midrule
            \bench-HuBERT                 & 95M   & \ding{55}  & 41.6 \\
            \bench-HuBERT                 & 95M   & \ding{51}  & 33.9 \\
            \bench-HuBERT (w. df) & 95M   & \ding{55}  & 41.0  \\
            \bench-HuBERT (w. df) & 95M   & \ding{51}  & 33.6  \\
            \midrule
            \bench-Conformer                     & 110M   & \ding{55}  & 49.5 \\
            \bench-Conformer                     & 110M   & \ding{51}  & 43.0   \\
            \bench-Conformer (w. df) & 110M   & \ding{55}  & 48.4  \\
            \bench-Conformer (w. df) & 110M   & \ding{51}  & 41.9   \\
            \bottomrule
        \end{tabular}
\end{table}

\subsection{Results}
We compare the previously mentioned baseline systems to both Whisper~\cite{whisper} and MMS~\cite{mms} family of models. We consider Whisper using $39$M, $74$M, $244$M, $769$M, and $1.55$B parameter models. For MMS we consider models trained on $61$ languages and $1,107$ languages. All MMS models contain $1.5$B parameters. Notice, that although Whisper and MMS models are multi-lingual, both were optimized over significantly larger datasets. 

Table~\ref{tab:wer} presents the Word-Error-Rates (WER) results computed over the Fleurs~\cite{conneau2023fleurs} benchmark. When considering comparison to Whisper models the provided baseline systems reach comparable or superior performance up until model size of $769$M parameters. When scaling the model size to $1.55$B, Whisper provides better performance while being between $\sim15$ times bigger. Although providing worse performance, we believe such models and results are interesting and valuable to the community as these could be important for use cases where performance can be compromised over significantly smaller models~\cite{kim2020review}. When comparing to MMS, the Conformer model trained on \bench provides comparable performance while HuBERT was found to be superior. Notice, that both Conformer and HuBERT are significantly smaller than the MMS model ($\sim15$x smaller). When comparing HuBERT and Conformer, results suggest HuBERT provides superior performance ($8-9$ absolute points) with and without LM decoding. 

Next, we evaluate the effect of the data filtering. We train both HuBERT and Conformer models using the filtered version as presented in \Cref{sec:df}. Notice, under this setup, the SSL, and pretraining part of HuBERT were still optimized using the whole dataset, while we modify only the fine-tuning part to use the filtered version only. Results are presented in \Cref{tab:wer} (bottom rows). Results suggest that the data filtering provides a small improvement for both HuBERT and the Comformer models. This suggests that overall in our dataset there is enough signal from the weak supervision to construct a performing acoustic model, however, there is still room for improvement in data quality assessment. We hope the speech community will adopt and develop such techniques in future research.

%% file: 06_discussion.tex
\section{Conclusion \& future work}
\label{sec:con}
In this work, we present \bench, a weakly supervised dataset in the Hebrew language, aimed at improving the development of AI-based speech processing tools directly dedicated to Hebrew. To further enhance the development of speech processing tools for Hebrew, we additionally provide two baseline systems, a self-supervised one and a fully supervised ASR acoustic model. Both \bench and the pre-trained models are released under the CC BY $4.0$ license. We hope the community will adopt such datasets and baseline systems together with other efforts in the field to advance the automatic development of speech-processing tools in Hebrew.

For future work, we plan to extend this dataset and provide higher-quality annotations in the form of transcriptions, speaker annotations, etc. Additionally, we plan to provide a subset of high-fidelity recordings which will be used to develop systems for generative tasks such as text-to-speech and voice conversion in Hebrew. 

\newpara{Acknowledgements} This research work was supported by the Israel Innovation Authority, grant number 78563.

%% file: main.bbl
\begin{thebibliography}{10}
\providecommand{\url}[1]{#1}
\csname url@samestyle\endcsname
\providecommand{\newblock}{\relax}
\providecommand{\bibinfo}[2]{#2}
\providecommand{\BIBentrySTDinterwordspacing}{\spaceskip=0pt\relax}
\providecommand{\BIBentryALTinterwordstretchfactor}{4}
\providecommand{\BIBentryALTinterwordspacing}{\spaceskip=\fontdimen2\font plus
\BIBentryALTinterwordstretchfactor\fontdimen3\font minus \fontdimen4\font\relax}
\providecommand{\BIBforeignlanguage}[2]{{%
\expandafter\ifx\csname l@#1\endcsname\relax
\typeout{** WARNING: IEEEtran.bst: No hyphenation pattern has been}%
\typeout{** loaded for the language `#1'. Using the pattern for}%
\typeout{** the default language instead.}%
\else
\language=\csname l@#1\endcsname
\fi
#2}}
\providecommand{\BIBdecl}{\relax}
\BIBdecl

\bibitem{wav2vec}
A.~Baevski, Y.~Zhou, A.~Mohamed, and M.~Auli, ``wav2vec 2.0: A framework for self-supervised learning of speech representations,'' \emph{Advances in neural information processing systems}, vol.~33, pp. 12\,449--12\,460, 2020.

\bibitem{audiopalm}
P.~K. Rubenstein, C.~Asawaroengchai, D.~D. Nguyen, A.~Bapna, Z.~Borsos, F.~d.~C. Quitry, P.~Chen, D.~E. Badawy, W.~Han, E.~Kharitonov \emph{et~al.}, ``Audiopalm: A large language model that can speak and listen,'' \emph{arXiv preprint arXiv:2306.12925}, 2023.

\bibitem{wang2021unispeech}
C.~Wang, Y.~Wu, Y.~Qian, K.~Kumatani, S.~Liu, F.~Wei, M.~Zeng, and X.~Huang, ``Unispeech: Unified speech representation learning with labeled and unlabeled data,'' in \emph{International Conference on Machine Learning}.\hskip 1em plus 0.5em minus 0.4em\relax PMLR, 2021, pp. 10\,937--10\,947.

\bibitem{librispeech}
V.~Panayotov, G.~Chen, D.~Povey, and S.~Khudanpur, ``Librispeech: an asr corpus based on public domain audio books,'' in \emph{2015 IEEE international conference on acoustics, speech and signal processing (ICASSP)}.\hskip 1em plus 0.5em minus 0.4em\relax IEEE, 2015, pp. 5206--5210.

\bibitem{librilight}
J.~Kahn, M.~Rivi{\`e}re, W.~Zheng, E.~Kharitonov, Q.~Xu, P.-E. Mazar{\'e}, J.~Karadayi, V.~Liptchinsky, R.~Collobert, C.~Fuegen \emph{et~al.}, ``Libri-light: A benchmark for asr with limited or no supervision,'' in \emph{ICASSP 2020-2020 IEEE International Conference on Acoustics, Speech and Signal Processing (ICASSP)}.\hskip 1em plus 0.5em minus 0.4em\relax IEEE, 2020, pp. 7669--7673.

\bibitem{voxpopuli}
C.~Wang, M.~Riviere, A.~Lee, A.~Wu, C.~Talnikar, D.~Haziza, M.~Williamson, J.~Pino, and E.~Dupoux, ``Voxpopuli: A large-scale multilingual speech corpus for representation learning, semi-supervised learning and interpretation,'' \emph{arXiv preprint arXiv:2101.00390}, 2021.

\bibitem{whisper}
A.~Radford, J.~W. Kim, T.~Xu, G.~Brockman, C.~McLeavey, and I.~Sutskever, ``Robust speech recognition via large-scale weak supervision,'' in \emph{International Conference on Machine Learning}.\hskip 1em plus 0.5em minus 0.4em\relax PMLR, 2023, pp. 28\,492--28\,518.

\bibitem{mms}
V.~Pratap, A.~Tjandra, B.~Shi, P.~Tomasello, A.~Babu, S.~Kundu, A.~Elkahky, Z.~Ni, A.~Vyas, M.~Fazel-Zarandi \emph{et~al.}, ``Scaling speech technology to 1,000+ languages,'' \emph{arXiv preprint arXiv:2305.13516}, 2023.

\bibitem{googleusm}
Y.~Zhang, W.~Han, J.~Qin, Y.~Wang, A.~Bapna, Z.~Chen, N.~Chen, B.~Li, V.~Axelrod, G.~Wang \emph{et~al.}, ``Google usm: Scaling automatic speech recognition beyond 100 languages,'' \emph{arXiv preprint arXiv:2303.01037}, 2023.

\bibitem{valle}
C.~Wang, S.~Chen, Y.~Wu, Z.~Zhang, L.~Zhou, S.~Liu, Z.~Chen, Y.~Liu, H.~Wang, J.~Li \emph{et~al.}, ``Neural codec language models are zero-shot text to speech synthesizers,'' \emph{arXiv preprint arXiv:2301.02111}, 2023.

\bibitem{voicebox}
M.~Le, A.~Vyas, B.~Shi, B.~Karrer, L.~Sari, R.~Moritz, M.~Williamson, V.~Manohar, Y.~Adi, J.~Mahadeokar \emph{et~al.}, ``Voicebox: Text-guided multilingual universal speech generation at scale,'' \emph{Advances in neural information processing systems}, vol.~36, 2024.

\bibitem{dns}
C.~K. Reddy, H.~Dubey, V.~Gopal, R.~Cutler, S.~Braun, H.~Gamper, R.~Aichner, and S.~Srinivasan, ``Icassp 2021 deep noise suppression challenge,'' in \emph{ICASSP 2021-2021 IEEE International Conference on Acoustics, Speech and Signal Processing (ICASSP)}.\hskip 1em plus 0.5em minus 0.4em\relax IEEE, 2021, pp. 6623--6627.

\bibitem{denoiser}
A.~Defossez, G.~Synnaeve, and Y.~Adi, ``Real time speech enhancement in the waveform domain,'' \emph{arXiv preprint arXiv:2006.12847}, 2020.

\bibitem{ssl_shua}
Y.~Dissen, F.~Kreuk, and J.~Keshet, ``{Self-supervised Speaker Diarization},'' in \emph{Proc. Interspeech 2022}, 2022, pp. 4013--4017.

\bibitem{yadav2022survey}
H.~Yadav and S.~Sitaram, ``A survey of multilingual models for automatic speech recognition,'' \emph{arXiv preprint arXiv:2202.12576}, 2022.

\bibitem{likhomanenko2020rethinking}
T.~Likhomanenko, Q.~Xu, V.~Pratap, P.~Tomasello, J.~Kahn, G.~Avidov, R.~Collobert, and G.~Synnaeve, ``Rethinking evaluation in asr: Are our models robust enough?'' \emph{arXiv preprint arXiv:2010.11745}, 2020.

\bibitem{campbell2008ethnologue}
L.~Campbell, ``Ethnologue: Languages of the world,'' 2008.

\bibitem{petrov2024language}
A.~Petrov, E.~La~Malfa, P.~Torr, and A.~Bibi, ``Language model tokenizers introduce unfairness between languages,'' \emph{Advances in Neural Information Processing Systems}, vol.~36, 2024.

\bibitem{black2019cmu}
A.~W. Black, ``Cmu wilderness multilingual speech dataset,'' in \emph{ICASSP 2019-2019 IEEE International Conference on Acoustics, Speech and Signal Processing (ICASSP)}.\hskip 1em plus 0.5em minus 0.4em\relax IEEE, 2019, pp. 5971--5975.

\bibitem{pratap2020mls}
V.~Pratap, Q.~Xu, A.~Sriram, G.~Synnaeve, and R.~Collobert, ``Mls: A large-scale multilingual dataset for speech research,'' \emph{arXiv preprint arXiv:2012.03411}, 2020.

\bibitem{conneau2023fleurs}
A.~Conneau, M.~Ma, S.~Khanuja, Y.~Zhang, V.~Axelrod, S.~Dalmia, J.~Riesa, C.~Rivera, and A.~Bapna, ``Fleurs: Few-shot learning evaluation of universal representations of speech,'' in \emph{2022 IEEE Spoken Language Technology Workshop (SLT)}.\hskip 1em plus 0.5em minus 0.4em\relax IEEE, 2023, pp. 798--805.

\bibitem{izre2001designing}
S.~Izre'el, B.~Hary, and G.~Rahav, ``Designing cosih: the corpus of spoken israeli hebrew,'' \emph{International Journal of Corpus Linguistics}, vol.~6, no.~2, pp. 171--197, 2001.

\bibitem{azogui2016open}
J.~Azogui, A.~Lerner, and V.~Silber-Varod, ``The open university of israel map task corpus (matacop),'' 2016.

\bibitem{marmorstein2022huji}
M.~Marmorstein and N.~Matalon, ``The huji corpus of spoken hebrew: An interaction-oriented design of a corpus,'' 2022.

\bibitem{sharoni2023saspeech}
O.~Sharoni, R.~Shenberg, and E.~Cooper, ``Saspeech: A hebrew single speaker dataset for text to speech and voice conversion,'' in \emph{Proc. Interspeech}, 2023.

\bibitem{marmor2023ivrit}
Y.~Marmor, K.~Misgav, and Y.~Lifshitz, ``ivrit. ai: A comprehensive dataset of hebrew speech for ai research and development,'' \emph{arXiv preprint arXiv:2307.08720}, 2023.

\bibitem{creativecreative}
C.~Commons, ``Creative commons attribution 4.0 international public license.''

\bibitem{silero_vad}
S.~Team, ``Silero vad: pre-trained enterprise-grade voice activity detector (vad), number detector and language classifier,'' 2021.

\bibitem{sohn1999statistical}
J.~Sohn, N.~S. Kim, and W.~Sung, ``A statistical model-based voice activity detection,'' \emph{IEEE signal processing letters}, vol.~6, no.~1, pp. 1--3, 1999.

\bibitem{pratap2023scaling}
V.~Pratap, A.~Tjandra, B.~Shi, P.~Tomasello, A.~Babu, S.~Kundu, A.~Elkahky, Z.~Ni, A.~Vyas, M.~Fazel-Zarandi \emph{et~al.}, ``Scaling speech technology to 1,000+ languages,'' \emph{arXiv preprint arXiv:2305.13516}, 2023.

\bibitem{hsu2021hubert}
W.-N. Hsu, B.~Bolte, Y.-H.~H. Tsai, K.~Lakhotia, R.~Salakhutdinov, and A.~Mohamed, ``Hubert: Self-supervised speech representation learning by masked prediction of hidden units,'' \emph{IEEE/ACM Transactions on Audio, Speech, and Language Processing}, vol.~29, pp. 3451--3460, 2021.

\bibitem{gulati2020conformer}
A.~Gulati, J.~Qin, C.-C. Chiu, N.~Parmar, Y.~Zhang, J.~Yu, W.~Han, S.~Wang, Z.~Zhang, Y.~Wu \emph{et~al.}, ``Conformer: Convolution-augmented transformer for speech recognition,'' \emph{arXiv preprint arXiv:2005.08100}, 2020.

\bibitem{ott2019fairseq}
M.~Ott, S.~Edunov, A.~Baevski, A.~Fan, S.~Gross, N.~Ng, D.~Grangier, and M.~Auli, ``fairseq: A fast, extensible toolkit for sequence modeling,'' \emph{arXiv preprint arXiv:1904.01038}, 2019.

\bibitem{graves2006connectionist}
A.~Graves, S.~Fern{\'a}ndez, F.~Gomez, and J.~Schmidhuber, ``Connectionist temporal classification: labelling unsegmented sequence data with recurrent neural networks,'' in \emph{Proceedings of the 23rd international conference on Machine learning}, 2006, pp. 369--376.

\bibitem{gulati20_interspeech}
A.~Gulati, J.~Qin, C.-C. Chiu, N.~Parmar, Y.~Zhang, J.~Yu, W.~Han, S.~Wang, Z.~Zhang, Y.~Wu, and R.~Pang, ``{Conformer: Convolution-augmented Transformer for Speech Recognition},'' in \emph{Proc. Interspeech 2020}, 2020, pp. 5036--5040.

\bibitem{vaswani2017attention}
A.~Vaswani, N.~Shazeer, N.~Parmar, J.~Uszkoreit, L.~Jones, A.~N. Gomez, {\L}.~Kaiser, and I.~Polosukhin, ``Attention is all you need,'' \emph{Advances in neural information processing systems}, vol.~30, 2017.

\bibitem{kim2020review}
C.~Kim, D.~Gowda, D.~Lee, J.~Kim, A.~Kumar, S.~Kim, A.~Garg, and C.~Han, ``A review of on-device fully neural end-to-end automatic speech recognition algorithms,'' in \emph{2020 54th Asilomar Conference on Signals, Systems, and Computers}.\hskip 1em plus 0.5em minus 0.4em\relax IEEE, 2020, pp. 277--283.

\end{thebibliography}
